\documentclass[a4paper,fleqn]{cas-sc}

\usepackage[authoryear]{natbib}

\def\tsc#1{\csdef{#1}{\textsc{\lowercase{#1}}\xspace}}
\tsc{WGM}
\tsc{QE}
\tsc{EP}
\tsc{PMS}
\tsc{BEC}
\tsc{DE}

\begin{document}
\let\WriteBookmarks\relax
\def\floatpagepagefraction{1}
\def\textpagefraction{.001}
\shorttitle{Maximum Matching Accuracy}
\shortauthors{K. Stillwagon et~al.}

\title [mode = title]{Maximum Matching Accuracy: An Instance Segmentation Evaluation Metric Utilizing Globally Optimal Matching}                      
\tnotemark[1]

\tnotetext[1]{Official Implementation: \url{https://github.com/kadenstillwagon/MMA}}

\author[1]{Kaden Stillwagon}[type=editor,
                        orcid=0009-0009-7081-2630]
\cormark[1]
\ead{kstillwagon26@gatech.edu}

\credit{Writing - original draft, Writing - review and editing, Methodology, Conceptualization, Software}

\author[2, 3]{Alexandra D. VandeLoo}[
    orcid=0009-0001-3035-3427]

\credit{Writing - review and editing}

\author[1, 3, 4]{Craig R. Forest}[
    orcid=0000-0001-5343-1769]
\cormark[2]

\credit{Writing - review and editing, Supervision, Funding acquisition}

\affiliation[1]{organization={College of Computing, Georgia Institute of Technology},
                city={Atlanta},
                postcode={30332}, 
                state={Georgia},
                country={United States}}

\affiliation[2]{organization={School of Materials Science and Engineering, Georgia Institute of Technology},
                city={Atlanta},
                postcode={30332}, 
                state={Georgia},
                country={United States}}

\affiliation[3]{organization={Wallace H. Coulter Department of Biomedical Engineering, Georgia Institute of Technology},
                city={Atlanta},
                postcode={30332}, 
                state={Georgia},
                country={United States}}

\affiliation[4]{organization={George W. Woodruff School of Mechanical Engineering, Georgia Institute of Technology},
                city={Atlanta},
                postcode={30332}, 
                state={Georgia},
                country={United States}}

\cortext[cor1]{Corresponding author}
\cortext[cor2]{Principal corresponding author}

\begin{abstract}
Reliable evaluation of instance segmentation models requires metrics that accurately and consistently reflect segmentation quality. However, the metrics most widely used in biological imaging carry fundamental mathematical weaknesses: hard Intersection-over-Union (IoU) thresholds that produce discontinuous, low sensitivity scoring; per-object normalization that distorts scores under object size variation; and greedy or one-to-many matching procedures that yield non-optimal, order-dependent correspondences. Together, these properties produce unintuitive and unreliable model rankings under common failure modes such as split cells, merged cells, and cell boundary imprecision. We propose Maximum Matching Accuracy (MMA), a threshold-free continuous metric that finds a globally optimal one-to-one matching between predicted and ground truth objects and aggregates total overlap using per-pixel normalization. We evaluate MMA against AP@50, PQ, SEG, and AJI across three experiments: synthetic failure cases, progressive corruption tests, and a model ranking comparison. MMA produces scores that are more stable, more sensitive, and more interpretable than existing alternatives, providing a principled foundation for fair instance segmentation benchmarking in biological cell imaging.
\end{abstract}


\maketitle

\section{Introduction}

Instance segmentation is the task of not only recognizing what objects are in an image, but also identifying the exact pixels belonging to each individual object. In cell biology, instance segmentation is particularly challenging and important for tasks ranging from medical diagnosis \citep{wu_single-cell_2020, mousavikhamene_morphological_2021, reta_segmentation_2015} to cell culture monitoring and analysis (\citep{way_predicting_2021, cole_light_2025, welter_machine_2024, he_morphology-based_2024, oja_automated_2018, kamat_single-cell_2024, vasilevich_correlation_2020, nassiri_systematic_2018}). Due to the difficulty of manually creating these segmentations of biological cells and the scale required to accomplish these objectives, wherein even a single image can have hundreds or thousands of cells, machine learning has become the dominant method used to automate segmentation. As a result, a growing subfield of computer vision has emerged to develop more accurate and robust cell segmentation models.

Advancing this field requires fair and accurate comparison between model outputs so that the advantages of one method over another can be determined. At the core of achieving fair model comparison is the creation and use of evaluation metrics for instance segmentation that assess model performance in a mathematically robust manner that aligns with expert intuition. However, several recent studies have identified shortcomings in widely adopted instance segmentation metrics, including limited sensitivity to segmentation quality, undesirable matching behavior, and ranking inconsistencies across competing methods \citep{hirling_segmentation_2024, chen_sortedap_2023, foucart_panoptic_2023, jena_beyond_2023, cheng_boundary_2021, karmakar_softpq_2025}. Inaccurate or biased metrics can result in unfair comparisons that cause the field to adopt an inferior method, diverting genuine progress. In biological imaging, the consequences of such errors extend beyond benchmarking: a model selected on the basis of a misleading metric may underperform in downstream tasks such as morphological profiling, cell counting, or clinical decision support, where segmentation quality directly affects scientific or diagnostic conclusions.

Standard segmentation evaluation metrics such as Panoptic Quality (PQ) \citep{kirillov_panoptic_2019}, mean Average Precision (mAP) \citep{everingham_pascal_2010}, and Intersection over Union (IoU) \citep{everingham_pascal_2010} are widely used across many domains. However, due to the application-driven nature of biological imaging work, a plethora of new accuracy metrics have been developed to fit specific needs. This significantly complicates comparison between methods and has also led to the misinterpretation of standard instance segmentation metrics. Hirling et. al. \citep{hirling_segmentation_2024} explores this issue in depth, finding six different interpretations of, for example, Average Precision (AP) and five different interpretations of mAP in biological image segmentation. Even more significantly, these researchers show that metric misinterpretations can reorder leaderboards in competitions that are highly influential to the field such as the 2018 Kaggle Data Science Bowl \citep{caicedo_nucleus_2019}, the 2021 Kaggle Sartorius Cell Instance Segmentation \citep{howard_sartorius_2021}, and the 2021 MIDOG (Mitosis Domain Generalization) \citep{aubreville_mitosis_2023}. This directly demonstrates how ill-defined instance segmentation evaluation metrics can cause improper comparison between models, potentially leading to use of and development upon inferior methods.

Beyond misinterpretation, the metrics themselves carry fundamental mathematical weaknesses that can produce unreliable model rankings. These weaknesses fall into three categories. First, many widely-used metrics rely on hard IoU thresholds to determine whether a predicted mask counts as a match: AP and PQ categorize each prediction as either true positive or false positive based on whether its IoU with a ground truth (GT) mask crosses a threshold (e.g. 0.5), while the Segmentation Accuracy Metric (SEG) from the Cell Tracking Challenge \citep{maska_benchmark_2014} uses an analogous overlap ratio threshold. This behavior creates discontinuities where marginally different predictions receive drastically different scores, and causes metrics to collapse to zero despite substantial overlap between predicted and GT masks. Second, metrics that normalize on a per-object basis can produce scores that are disproportionately distorted by small objects, obscuring the true quality of segmentation over the majority of the image. Third, the matching procedures used to assign correspondences between predicted and GT masks are often either greedy, and therefore non-optimal, or permit one-to-many assignments in which a single predicted mask can simultaneously satisfy multiple GT objects. These behaviors can introduce ambiguity and order dependence into the final score. Collectively, these design choices mean that under common biological segmentation failure modes (e.g. split cells, merged cells, or cell boundary imprecision) existing metrics can produce rankings that do not reflect true segmentation quality. A metric that avoids each of these failure modes is needed to enable reliable model comparison.

We propose Maximum Matching Accuracy (MMA) as a novel instance segmentation metric. MMA is a threshold-free, continuous metric that addresses each of these shortcomings through three core design choices: globally optimal one-to-one matching, continuous overlap scoring, and pixel-level global normalization. To establish our baselines, we performed a unified characterization of four widely-used segmentation metrics, AP@50, PQ, SEG, and AJI, analyzing their mathematical properties and failure modes within a common framework. We evaluate MMA against these metrics as well as an ablated version of MMA we term “MMA-Greedy” across two complementary experiments: synthetic failure cases that isolate known metric weaknesses and progressive corruption tests that measure stability and sensitivity under realistic segmentation degradation. Across both settings, MMA produces scores that are more stable and more sensitive than existing alternatives, establishing it as a principled foundation for fair instance segmentation evaluation in biological cell imaging. Importantly, we additionally evaluate the impact of MMA in ranking segmentation models, finding that MMA reports a different top model than baseline metrics in up to 50\% of evaluated cases. While we present MMA in the context of biological cell image segmentation, we suggest, but have not shown, that its use may be extended to all domains involving instance segmentation that disallows overlap between predicted segmentations. This remains a promising area for future work.

\section{Related Work}

We categorize existing instance segmentation evaluation metrics into three broad classes: detection-style metrics, continuous overlap metrics, and hybrid metrics that combine properties of both approaches. Detection-style metrics evaluate segmentation quality through thresholded matching and categorical counting of true positives (TP), false positives (FP), and false negatives (FN). Continuous overlap metrics instead measure segmentation quality directly through pixel-level overlap and are particularly common in biological cell segmentation tasks where sensitivity to small boundary changes is important. Hybrid metrics attempt to balance these approaches by combining thresholded instance matching with continuous overlap evaluation.

\subsection{Detection-style Metrics}
The detection-style category of segmentation evaluation metrics function by grouping predicted and GT segmentations into three categories (TP, FP, and FN) and compute an Average Precision (AP) score. In labeling these categories, an IoU threshold is specified and any prediction with an IoU score greater than this threshold is considered a TP, any prediction with a lower IoU is a FP, and all GT objects without corresponding predicted masks over this threshold are FN. AP-style IoU-threshold metrics were originally popularized in object detection benchmarks such as PASCAL VOC \citep{everingham_pascal_2010} and COCO \citep{lin_microsoft_2015}, where AP is calculated as the area under the precision-recall curve specified in Equation \ref{eq:classic_ap} in which $r$ is recall and $p(r)$ is the associated precision as follows: 
\begin{equation}
  AP_1 = \int p(r) dr
  \label{eq:classic_ap}
\end{equation}

This metric became widely adopted in biological instance segmentation following challenges such as the 2018 Kaggle Data Science Bowl \citep{caicedo_nucleus_2019}. However, many of these works reinterpreted the AP metric with different mathematical formulations \citep{hirling_segmentation_2024}. The 2018 Kaggle Data Science Bowl defined average precision as the mean precision value at IoU thresholds $t$ from 0.5 to 0.95 with a step size of 0.05 as follows:
\begin{equation}
  mAP = \frac{1}{|thresholds|} \sum_t \frac{TP(t)}{TP(t) + FP(t) + FN(t)}
  \label{eq:kaggle_2018_mAP}
\end{equation}
Many current biological segmentation works \citep{stringer_cellpose_2021, pachitariu_cellpose-sam_2025, marks_cellsam_2025} evaluate and compare their models using a simplified version of this metric that uses only the precision value at 0.5 IoU. We refer to this metric as AP@50, following previous works \citep{pachitariu_cellpose-sam_2025, marks_cellsam_2025}. Detection-style metrics introduce discontinuities and insensitivity around arbitrarily-chosen thresholds, resulting in both instability and a lack of precision. 

\subsection{Continuous Overlap Metrics}
The shortfalls of categorical detection-style metrics motivate the desire for continuous, per-pixel metrics with more sensitive scores. Two metrics: segmentation accuracy measure (SEG) \citep{maska_benchmark_2014} and Aggregated Jaccard Index (AJI) \citep{kumar_dataset_2017}, emerged as prominent methods for continuous assessment of segmentation quality and model performance comparison. SEG, introduced by the Cell Tracking Challenge, computes the mean Jaccard similarity index ($J$) \citep{jaccard_etude_1901} of matched reference and segmented objects. A reference object $R$ and segmented object $S$ are considered matched if 
\begin{equation}
|R \cap S| > 0.5 |R|
  \label{eq:SEG_match}
\end{equation}
and the Jaccard similarity is defined as:
\begin{equation}
  J(R, S) = \frac{|R \cap S|}{|R \cup S|}
  \label{eq:SEG}
\end{equation}

In SEG, reference sets $R$ that don’t meet the matching condition for any segmented object $S$ are matched to the empty set. This property not only makes SEG discontinuous at the threshold, but also allows a single segmented object $S$ to be matched to multiple reference objects $R$. In contrast, AJI greedily finds the segmented mask of highest IoU $S_i^*$ for each GT object $G_i$ and adds their intersection to the numerator and their union to the denominator with the magnitudes of any unused predicted segmentations $S_j^0$ also added to the denominator as shown below:
\begin{equation}
  AJI = \frac{\sum_i |G_i \cap S_i^*|}{\sum_i |G_i \cup S_i^*| + \sum_j |S_j^0|}
  \label{eq:AJI}
\end{equation}
While AJI doesn’t use arbitrary IoU thresholding like SEG or Detection-style metrics, its matching method is greedy, and therefore non-optimal, and also allows for a segmented object to be matched to many different GT objects. 

\subsection{Hybrid Metrics}
Hybrid metrics exist at the intersection between detection-based and continuous overlap metrics and were developed to retain the strengths and minimize the weaknesses of each of these methods. A key example of this is Panoptic Quality (PQ) \cite{kirillov_panoptic_2019}, first introduced for panoptic segmentation and adapted for use in many instance cell segmentation works \citep{graham_hover-net_2019, horst_cellvit_2023, verma_monusac2020_2021, graham_conic_2021}. Panoptic quality measures segmentation quality (SQ) and recognition quality (RQ) simultaneously as shown below where p is a predicted object, g is a GT object, and (p, g) is a matched pair:
\begin{equation}
\mathrm{PQ} =
    \underbrace{
        \frac{\sum_{(p,g)\in TP} \mathrm{IoU}(p,g)}
        {|TP|}
    }_{\text{segmentation quality (SQ)}}
    \times
    \underbrace{
        \frac{|TP|}
        {|TP| + \frac{1}{2}|FP| + \frac{1}{2}|FN|}
    }_{\text{recognition quality (RQ)}}
\label{eq:PQ}
\end{equation}
The segmentation quality represents the mean IoU between TP pairs and recognition quality represents a measure similar to average precision, often called F1-score \citep{everingham_pascal_2010}. TP pairs are matched by having IoU strictly greater than 0.5 as in AP@50. While this metric captures both segmentation and representation quality, it still relies upon thresholding, causing instability around those threshold values.

\section{Methods}

In this work, we attempt to address the limitations of each of the above baseline metrics through MMA: a threshold-free, continuous instance segmentation metric with optimal one-to-one matching. We analyse the stability and logic of AP@50, SEG, AJI, and PQ against MMA directly to determine whether MMA enables more fair and accurate cell segmentation model comparison.  

\subsection{Maximum Matching Accuracy (MMA)}
\subsubsection{MMA Matching Procedure}
Our proposed Maximum Matching Accuracy metric enforces a strict one-to-one correspondence between GT and predicted segmentation masks. Specifically, MMA finds the globally optimal one-to-one matching between GT and predicted masks by maximizing the total area of intersection between matched instances. This is accomplished by formulating a maximum bipartite matching problem on a graph representing the instance-matching situation. The graph is constructed as follows:
\begin{itemize}
    \item $\textit{Nodes}$: one node is created for each GT mask and for each model predicted mask.
    \item $\textit{Edges}$: an edge is built between any pair of nodes representing a GT and a model predicted mask that overlap. The edge weight is determined by the cardinality of the overlap, measured in pixels.
    \item $\textit{Key Property}$: this is a bipartite graph with disjoint sets representing GT and model predicted nodes.
\end{itemize}

\begin{figure}
	\centering
	\includegraphics[width=.9\textwidth]{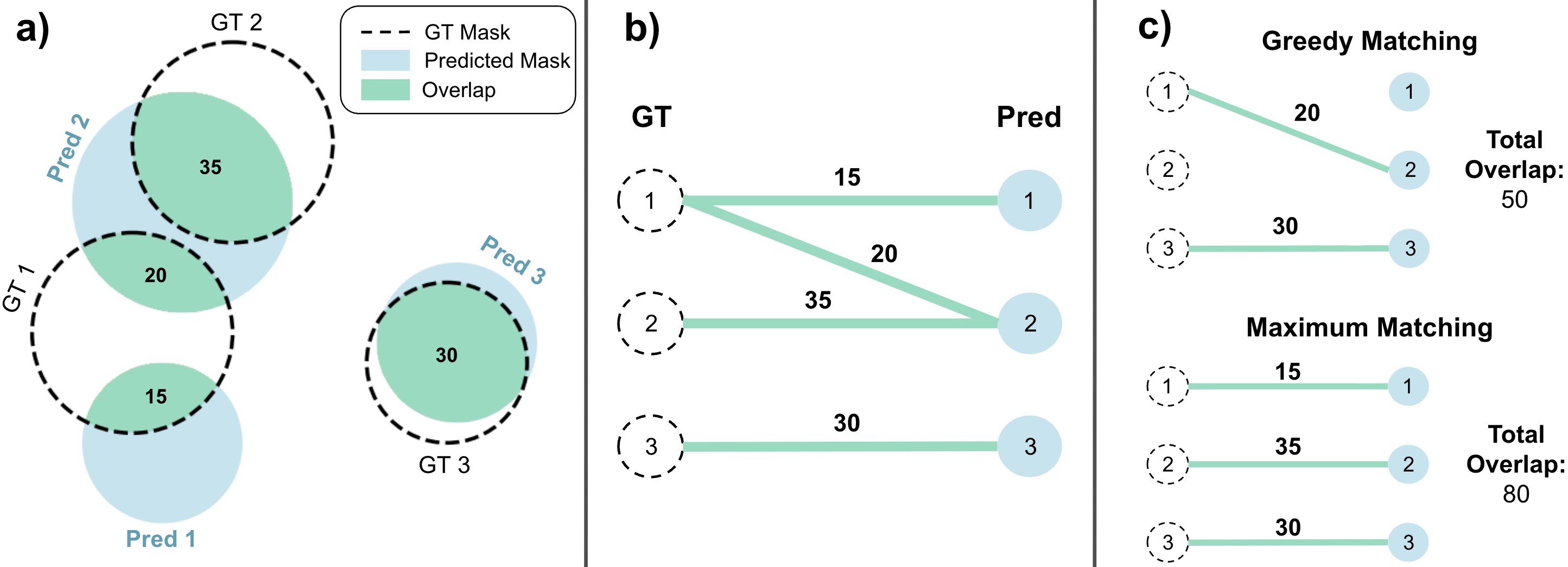}
	\caption{\textbf{Maximum Bipartite Matching for Instance Segmentation Assignment.} \textbf{(a)} Example ground truth (GT) and predicted (Pred) segmentation masks. Areas of overlap are represented by numbers in regions where GT and Pred overlap. \textbf{(b)} A bipartite graph representing the segmentation scenario in (a) including nodes for GT masks on the left, nodes for predicted masks on the right, and edges between nodes that overlap with weights determined by the area of overlap. \textbf{(c)} Matching results after solving Greedy Matching and Maximum Bipartite Matching on the graph in (b). Maximum Bipartite Matching finds an optimal solution that maximizes total overlap area between GT and predicted masks, while Greedy Matching finds a locally optimal solution that prevents global optimality.}
	\label{FIG:max_bipartite_matching_example}
\end{figure}

Fig. \ref{FIG:max_bipartite_matching_example}(a, b) shows an example of how a set of GT and predicted segmentation masks are converted into a bipartite graph. We use the max\_weight\_matching function from the NetworkX library \citep{hagberg_proceedings_2008} to solve the maximum bipartite matching problem on the constructed graph. This function returns the matching (a set of edges in which no node occurs more than once) with the maximum sum of edge weights as seen in Fig. \ref{FIG:max_bipartite_matching_example}(c). It has a time complexity of O($n^3$), where n is the number of nodes. The edges contained in the matching specify the one-to-one association between GT and predicted segmentation masks with maximum total overlap. We use this association in the calculation of MMA.

\subsubsection{MMA Calculation}
Maximum Matching Accuracy is calculated per-image through a simple fractional structure. To compute the numerator, we sum the total pixel area of each intersection between matched GT and predicted instances. The denominator is equal to the cardinality of the union over all ground-truth or model-labeled pixels and acts as a normalizing feature. We show the complete formulation of MMA below, where $GT_i$ and $P_i$ represent the ground truth and predicted instances in matched set $i$ and $GT$ and $P$ represent all ground truth and predicted pixels, respectively:
\begin{equation}
  MMA = \frac{\sum_i |GT_i \cap P_i|}{|GT \cup P|}
  \label{eq:MMA}
\end{equation}

\subsubsection{Properties of MMA}
MMA holds several key properties that influence its ability to accurately and reliably measure instance segmentation performance. The first of these is that the matching procedure used to compute MMA enforces one-to-one matching and is globally optimal. Concretely, there is no other matching between GT and predicted segmentations that has a higher total area of intersection and, therefore, no alternative matching that could lead to a higher MMA score. Secondly, MMA is a threshold-free metric that uses continuous overlap weights instead of categorical TP/FP/FN assignment. This allows for small changes in predicted masks to be accurately and proportionally reflected in the resulting score. Additionally, MMA symmetrically accounts for FP and FN failure modes, naturally balancing precision and recall. Finally, MMA normalizes an image globally so each labeled pixel holds the same weight and is counted exactly once. This is in contrast to per-object normalization where each object holds the same weight and pixels may be counted multiple times as TP, FP, or FN for different matchings. As a result, smaller objects hold less weight than large objects under the MMA formulation.

\subsubsection{MMA-Greedy}
To enable a direct evaluation of the maximum bipartite matching policy, we introduce a variant of MMA that we term MMA-Greedy. This metric has the same formulation as MMA (shown in Eq. \ref{eq:MMA}) and differs only in the manner that model predicted and GT objects are matched. Instead of finding a one-to-one matching that maximizes total overlap, MMA-Greedy uses a greedy one-to-one matching, where each GT object is considered in sequence and is matched to the unmatched predicted object that overlaps the most with it. An example of this matching is shown in Fig. \ref{FIG:max_bipartite_matching_example}. With this ablated metric, the direct impact of using a globally optimal matching over a greedily chosen one can be analyzed. We examine MMA-Greedy alongside MMA and all baseline metrics in each experiment.

\subsection{Evaluation and Comparison}
We compare MMA against baseline metrics over three experiments: synthetic failure cases, progressive corruption tests, and model ranking comparisons. 

\subsubsection{Synthetic Failure Cases}
We constructed a series of controlled synthetic instance segmentation examples designed to isolate specific behaviors of evaluation metrics under common failure modes. Each example consists of simple synthetic ground truth and predicted instance masks that are easily interpretable and highlight key differences between MMA and baseline metrics. The evaluated scenarios include threshold sensitivity, over-segmentation, under-segmentation, unequal object size weighting, greedy assignment ambiguity, and one-to-many matching.

These failure cases were designed to reflect biologically relevant segmentation errors commonly encountered in microscopy data, including split cells, merged cells, false detections, and ambiguous instance correspondences. For each scenario, we computed scores for MMA and all baseline metrics to evaluate how differences in thresholding, normalization, overlap aggregation, and matching methodology influence metric behavior. The experiments specifically evaluate several known properties of instance segmentation metrics, including sensitivity to arbitrary overlap thresholds, handling of FP predictions, weighting of differently sized objects, use of greedy assignment strategies, and allowance of one-to-many correspondences. By isolating these behaviors in controlled settings, the experiments provide interpretable examples of how metric design choices influence reported segmentation quality.

\subsubsection{Progressive Corruption Tests}
We evaluate the stability of each metric under various forms of progressive corruption. Eight types of realistic and biologically relevant corruption are tested, which are defined as follows: 
\begin{itemize}
    \item $\textit{Erosion}$: at each iteration the outermost layer of pixels is removed from each segmentation mask.
    \item $\textit{Dilation}$: at each iteration, every segmentation mask is dilated once using the cv2.dilate \citep{bradski_opencv_2000} function with a 3x3 kernel. If two masks overlap as a result of the dilation, the smaller mask is assigned the overlapping pixels.
    \item $\textit{Fragmentation}$: at each iteration, 30\% of the segmentation masks, chosen randomly, are fragmented by randomly choosing two new centers within the current mask and assigning pixels to two new masks based on which center they are closest to. Masks that have an area of less than 25 pixels are never chosen to be fragmented.
    \item $\textit{Clumping/Grouping}$: at each iteration 20\% of the segmentation masks, chosen randomly, are merged with the segmentation mask that they are closest to (based on the distance between their centroids). 
    \item $\textit{Removal}$: at each iteration 20\% of the remaining segmentation masks, chosen randomly are removed (e.g. set as background).
    \item $\textit{Addition}$: at each iteration 10 circular masks with radius’ randomly chosen from 6 to 20 pixels are added to random unlabeled locations in the image such that they do not extend over the image border or into other segmentations. If there is no location where a mask of the chosen radius can fit without overlapping with other masks or extending over the border, then a circle of maximum possible radius is added to the point with a maximum radius of unlabeled space surrounding it.
    \item $\textit{Variable Size Addition}$: at each iteration, 10 circular masks of pixel radius 4 times the iteration number are added to the image. If there is no location where a mask of the chosen radius can fit, then it is handled in the same manner as the Addition corruption type. Unlike previous corruption types, this form of corruption does not build off of previous iterations but is reset to the original GT masks before each iteration. In this way, the effect of the added masks’ size can be analyzed directly without considering the number of masks that have been added.
    \item $\textit{Spatial Shifting}$: at each iteration, 50\% of the segmentation masks are shifted in a randomly chosen direction with a magnitude equal to the iteration number. Similarly to the Dilation corruption type, smaller masks are assigned pixels that are overlapping with other masks. Additionally, like the Variable Size Addition corruption type, this form of corruption is reset to the original GT masks before each iteration.
\end{itemize}

We perform nine iterations of each corruption type on every image from the test set of the LIVECell dataset \citep{edlund_livecelllarge-scale_2021}. Scores are then computed for each metric on every image at all corruption stages and averaged over the dataset. We plot the mean values of each metric at every corruption stage and compare the stability and sensitivity of the metrics to all corruption types. 

\subsubsection{Model Ranking Comparison}
In a final experiment, we analyse how MMA changes rankings between three instance cell segmentation models on the LIVECell test set compared to baseline metrics. Specifically, we compute each metric on outputs from the DINOCell \citep{stillwagon_self-supervised_2026}, Cellpose-SAM \citep{pachitariu_cellpose-sam_2025}, and SAMCell \cite{vandeloo_samcell_2025} models. These metrics are used to rank the three models on each example independently. We then calculate two statistics for each baseline metric individually that quantify their disagreement with MMA’s ranking of model outputs. The first of these we term $\textit{pairwise disagreement}$ and calculate it as the fraction of model pairs whose relative ordering differs between MMA and metric $m$. To formalize this, let $M$ denote the set of evaluated models and $N$ denote the number of images in the LIVECell test set. For each image $k$, we define a ranking produced by metric $m$ as $r^{(k)}_m$. For every unordered pair of models ($i, j$), we determine whether MMA and metric $m$ agree on their relative ordering using the indicator function $1[*]$ and compute the mean over all images and pairs as follows:
\begin{equation}
  D_{pairwise} = \frac{1}{N} \sum_{k=1}^N \Big( \frac{1}{(\frac{|M|}{2})} \sum_{i < j} 1\big[(r^{(k)}_{MMA}(i) - r^{(k)}_{MMA}(j))(r^{(k)}_{m}(i) - r^{(k)}_{m}(j)) < 0 \big] \Big)
  \label{eq:pairwise_disagreement}
\end{equation}
The second measure we call $\textit{top-1 disagreement}$ and calculate it as the percentage of examples in which metric m selects a different best-performing model than MMA. Using the same notation as above, we define this measure as:
\begin{equation}
  D_{top1} = \frac{1}{N} \sum_{k=1}^N 1\big[argmin_i r^{(k)}_{MMA}(i) \neq argmin_i r^{(k)}_{m}(i) \big]
  \label{eq:top_1_disagreement}
\end{equation}
We calculate both of these measures for each baseline metric, enabling analysis on the impact of using MMA to rank the quality of segmentation model outputs.

\section{Results and Discussion}
\subsection{Synthetic Failure Cases}
\begin{figure}
	\centering
	\includegraphics[width=.61\textwidth]{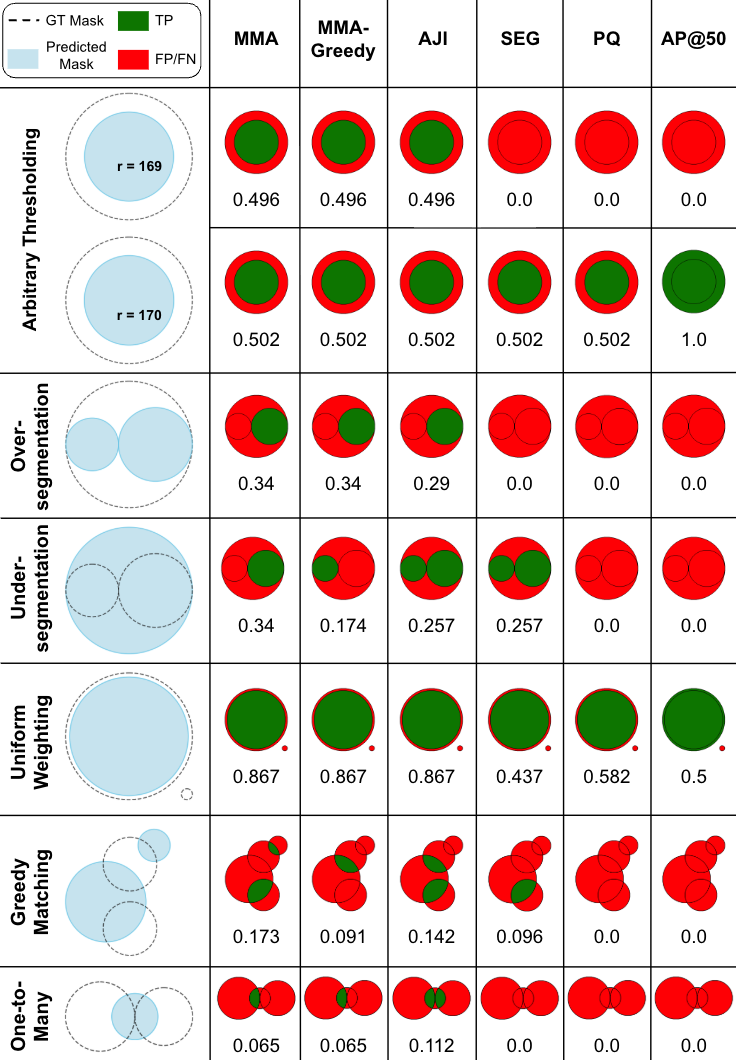}
	\caption{\textbf{MMA Demonstrates Robustness on Structured Test Cases where Other Metrics Behave Unintuitively.} MMA and five baseline segmentation quality metrics are evaluated on six structured scenarios representing common segmentation failure modes. In the left column are drawings of each scenario with filled blue circles representing predicted masks and dashed black circle outlines representing ground truth masks. To the right of these are depictions of how each metric interprets true positives (green) and false positives/negatives (red) in the scenarios along with their associated scores. Metrics relying on hard thresholds (SEG, PQ, and AP@50) exhibit discontinuous behavior and frequently collapse to zero despite substantial overlap. Baseline metrics additionally produce unintuitive penalties through per-object normalization (e.g. SEG in $\textit{Uniform Weighting}$) and multi-counting of pixels (e.g. AJI in $\textit{Over-Segmentation}$). Greedy matching procedures in MMA-Greedy and AJI can lead to non-optimal matchings and metrics that allow one-to-many matchings (AJI and SEG) can produce misleading results. In contrast, MMA produces sensitive and consistent scores through optimal one-to-one matching and global pixel-level normalization.}
	\label{FIG:synthetic_failure_cases}
\end{figure}

Existing instance segmentation metrics exhibit several limitations under realistic failure modes, often producing scores that do not align with the true quality of the predicted segmentations. Fig. \ref{FIG:synthetic_failure_cases} highlights several of these failure cases designed to expose how MMA and existing metrics differ in three core design choices: (1) thresholding strategy, (2) score normalization and weighting, and (3) matching methodology. Across these cases, MMA consistently produces scores that more closely reflect the actual quality of segmentation overlap by combining continuous overlap scoring, globally optimal one-to-one matching, and pixel-level normalization. In contrast, existing metrics often exhibit discontinuities, order dependence, or unintuitive penalization caused by a variety of design choices including: object-centric normalization, thresholding, and non-optimal matching rules.

\subsubsection{Thresholding Behavior Comparison}
A major difference between MMA and several existing metrics is the absence of binary thresholding. AP@50, PQ, and SEG all rely on binary matching decisions that introduce discontinuous behavior near arbitrary overlap thresholds. This effect is most evident in the $\textit{Arbitrary Thresholding}$ example, where a minimal increase in prediction radius (169px to 170px) causes abrupt transitions in AP@50, SEG, and PQ once their matching thresholds are exceeded. Specifically, AP@50 demonstrates its property of all-or-nothingness when it moves from 0 to 1.0 after IoU surpasses 0.5. Similar behavior can be seen in the $\textit{Over-segmentation}$, $\textit{Under-segmentation}$, $\textit{Greedy Matching}$, and $\textit{One-to-Many}$ examples where PQ and AP@50 are 0 despite substantial overlap between GT and predicted segmentation masks. In this way, these metrics are shown to be insensitive to partial correctness.

The $\textit{Under-segmentation}$ and $\textit{Greedy Matching}$ examples further illustrate that threshold definitions themselves can substantially alter metric behavior. While PQ and AP@50 are 0 in these examples, SEG assigns a nonzero score because its matching criterion is determined by the ratio between GT size and overlap size rather than IoU-based. As a result, the prediction is considered sufficiently matched despite failing the IoU threshold required by PQ and AP@50. With the SEG matching criterion, a predicted segmentation can be arbitrarily larger than a GT object and still be considered a match as long as a majority of the GT’s pixels are covered by it. IoU thresholding, in contrast, is much stricter, requiring the overlap to be at least half the size of the union between the predicted and GT masks. This demonstrates that metrics using different threshold definitions may produce fundamentally different evaluations for the same segmentation output.

These thresholding schemes create behavior in which nearly identical predictions can receive drastically different scores. In contrast, MMA, MMA-Greedy, and AJI vary smoothly with overlap quality, avoiding discontinuities caused by arbitrary thresholds. Additionally, these continuous metrics preserve partial credit, allowing the score to degrade gradually with segmentation quality rather than collapsing to binary success or failure.

\subsubsection{Normalization and Weighting Comparison}
The experiments also reveal substantial differences between object-centric and pixel-centric normalization strategies. MMA uses global pixel-level normalization, where each pixel contributes exactly once to the final score. In contrast, AJI, SEG, PQ, and AP@50 rely on object-centric formulations that either normalize overlap on a per-object basis before aggregation or permit implicit multi-counting of pixels across multiple matchings. 

This distinction becomes particularly important in the $\textit{Uniform Weighting}$ example. SEG, PQ, and AP@50 assign disproportionately low scores because each GT object contributes equally to the final metric regardless of size. Consequently, a very small unmatched object penalizes the score as much as the large, nearly correctly segmented object. In contrast, MMA, MMA-Greedy, and AJI weight contributions proportionally to overlap area, causing the dominant large overlap to drive an appropriately high final score.

The $\textit{Under-segmentation}$ and $\textit{Greedy Matching}$ examples further expose limitations of object-centric normalization. These examples demonstrate scenarios in which pixels can be counted multiple times. In AJI and SEG, overlap is computed independently for each matched pair: overlapping pixels contribute to both the numerator and denominator, while non-overlapping pixels contribute to the denominator. As a result, when two GT objects are matched to a single predicted object, as is the case in $\textit{Greedy Matching}$ and $\textit{One-to-Many}$ for AJI and $\textit{Under-Segmentation}$ for both AJI and SEG, the pixels within the shared predicted object are counted as TPs for one of the matchings and FNs for the other. 

MMA avoids this ambiguity through globally consistent normalization, in which each pixel is classified only once. This produces scores that more directly reflect total overlap quality across the image rather than the average behavior of independently evaluated object pairs. Importantly, this property is what allows MMA to have higher scores than AJI and SEG in $\textit{Under-segmentation}$ and $\textit{Greedy Matching}$ despite matching smaller total pixel areas. 

However, this global formulation also introduces tradeoffs. In AJI any remaining unmatched predicted objects are fully added to the denominator. The $\textit{Over-Segmentation}$ example illustrates how this can lead to double-counting when an unmatched predicted object overlaps a GT object. In this case, the overlapping pixels contribute once as FNs within the matched GT pair and again as FPs because the redundant prediction remains unmatched. In contrast, MMA assigns the same score regardless of whether the smaller redundant prediction exists because unmatched fragments do not contribute separately to the globally normalized overlap. As a result, MMA can be less sensitive to certain redundant predictions than metrics that explicitly penalize unmatched objects independently.

\subsubsection{Matching Strategy Comparison}
Fig. [2] also demonstrates that matching methodology substantially influences metric behavior. Two matching strategies are present in the baseline metrics: SEG, PQ, and AP@50 use thresholding to determine matching and MMA-Greedy and AJI use a greedy approach on GT objects. MMA, instead, computes a globally optimal assignment between GT and predicted masks.

The $\textit{Greedy Matching}$ and $\textit{Under-segmentation}$ examples illustrate the limitations of greedy assignment. MMA reports a higher score than MMA-Greedy because it selects the globally optimal correspondence configuration. MMA-Greedy instead commits to an early locally favorable match that prevents a superior overall assignment later. A higher score is desirable here because MMA and MMA-Greedy differ only in their matching strategies, so a higher score directly indicates that MMA finds a better matching. Importantly, greedy matching behavior is order dependent: processing GT objects in a different order could produce a different final score and potentially recover the optimal solution. This reveals an instability inherent to greedy matching strategies.

The experiments additionally highlight differences between one-to-one and one-to-many matching. MMA enforces strict one-to-one correspondences, ensuring that each predicted and GT mask contribute to at most one match. In contrast, AJI and SEG allow one predicted object to match multiple GT objects. This behavior is evident in the $\textit{One-to-Many}$, $\textit{Under-Segmentation}$, and $\textit{Greedy Matching}$ examples, where multiple ground masks partially benefit from the same predicted region. One-to-many matching breaks the intuitive reasoning that each predicted object corresponds to at most one GT object and introduces ambiguity in score interpretation. Allowing multiple GT objects to match with a single predicted object can artificially increase reported correspondence quality, potentially leading to inaccurate rankings and comparisons. MMA avoids this issue through exclusive assignments that produce globally consistent overlap accounting.

Collectively, these experiments show that MMA differs from existing metrics in three fundamental ways: it removes threshold-induced discontinuities, uses strict pixel-level normalization, and computes globally optimal one-to-one assignments. Existing metrics often combine thresholding, object-centric averaging, and greedy or one-to-many matching strategies, which can produce unintuitive behavior under common segmentation failure modes such as over-segmentation, under-segmentation, and ambiguous correspondences. MMA instead provides a smoother and more globally coherent measure of segmentation quality that better reflects the total overlap between appropriately matched predictions and GT.

\subsection{Progressive Corruption Tests}
\begin{figure}
	\centering
	\includegraphics[width=.82\textwidth]{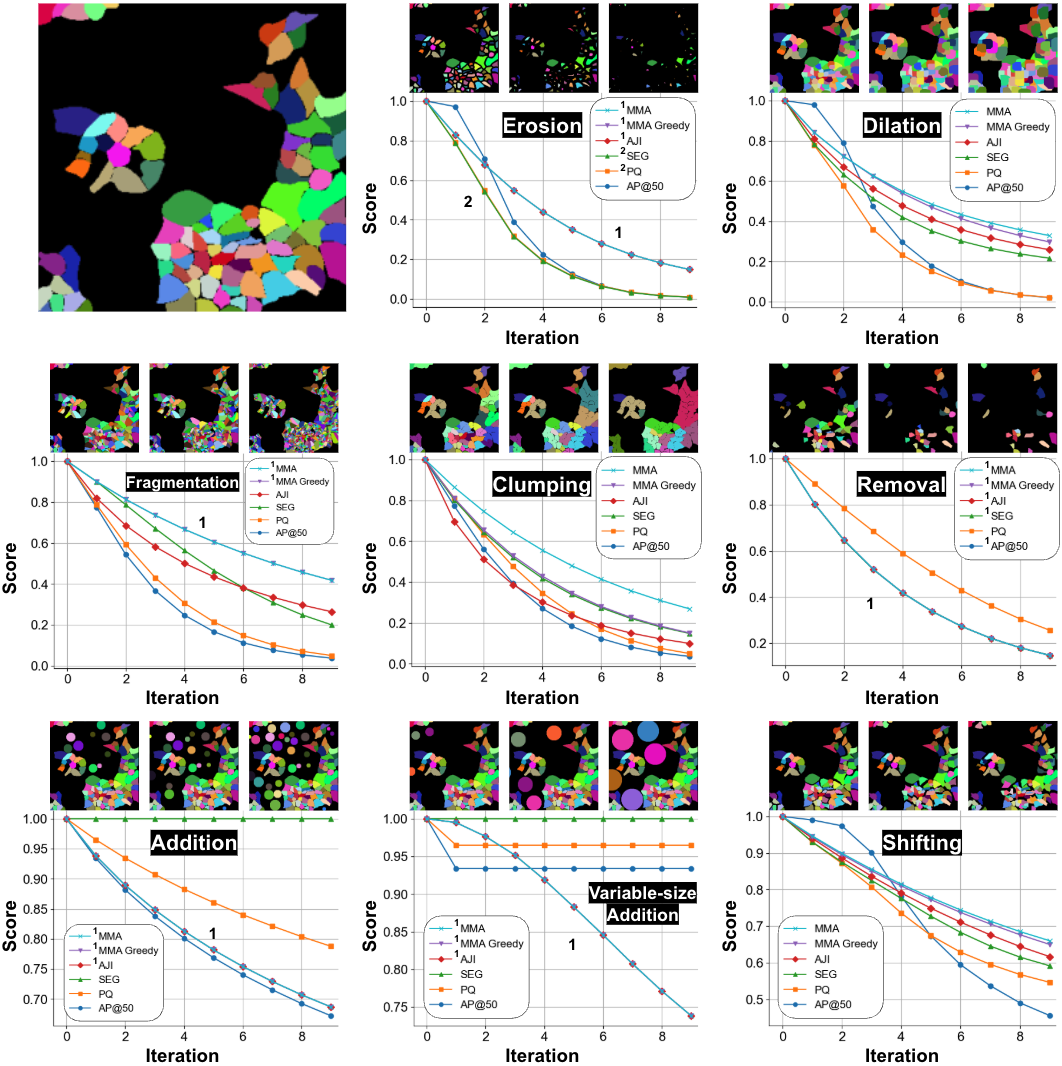}
	\caption{\textbf{MMA Consistently Exhibits Greater Stability and Sensitivity than Baseline Metrics Under Progressive Segmentation Corruption.} Ground truth instance masks were progressively degraded using eight controlled corruption operations representing common segmentation failure modes. Metric responses are shown as corruption severity increases across iterations. Additionally, a segmentation example from LIVECell \citep{edlund_livecelllarge-scale_2021} is shown at different iterations (1, 5, 9) for each corruption type with the original masks shown in the top left. Some graphs include numbers in their figures and legends to indicate metric curves that fully overlap. Threshold-based metrics (particularly PQ and AP@50) exhibit abrupt nonlinear degradation and early score collapse under erosion, dilation, fragmentation, and shifting due to hard matching thresholds. PQ and AP@50 fail to account for false positive (FP) magnitude and SEG fails to account for FPs in general as shown in the $\textit{Addition}$ and $\textit{Variable-Size Addition}$ experiments. In contrast, MMA produces smooth degradation curves across all corruption types while remaining responsive to both overlap deterioration and increasing FP magnitude. Compared to MMA-Greedy and AJI, MMA additionally maintains greater stability under ambiguous matching scenarios such as clumping and shifting through globally optimal assignment and per-pixel normalization. Collectively, these experiments demonstrate that MMA provides a more stable, sensitive, and interpretable measure of segmentation quality under progressive degradation.
}
	\label{FIG:progressive_corruption_experiment}
\end{figure}

To evaluate how segmentation metrics respond to progressive degradation in prediction quality, we systematically applied a series of controlled corruption operations to GT instance masks and measured the response of MMA and baseline metrics across increasing corruption severity. The corruptions were designed to isolate distinct failure modes commonly encountered in real-world instance segmentation tasks, including boundary distortions (erosion and dilation), over- and under-segmentation (fragmentation and clumping), localization errors (shifting), and object-level detection failures (removal and FP addition). An ideal segmentation metric should satisfy two key properties under these conditions: (1) stability, where scores change smoothly as segmentation quality gradually deteriorates, and (2) sensitivity, where structurally meaningful segmentation errors produce appropriately large score reductions. Fig. \ref{FIG:progressive_corruption_experiment} compares MMA against baseline segmentation metrics under these corruption settings, revealing substantial differences in thresholding behavior,  FP handling, overlap aggregation, and instance matching methodology.

\subsubsection{Thresholding Behavior Under Progressive Corruption}
Across most corruption types, threshold-based metrics exhibit rapid deterioration in early iterations as predictions begin to fall below their matching thresholds. This behavior is particularly evident in $\textit{Dilation}$ and $\textit{Fragmentation}$ for AP@50 and PQ and $\textit{Erosion}$ for AP@50, PQ, and SEG, where these metrics often collapse toward zero despite corrupted masks maintaining significant overlap with the original GT. These threshold-driven transitions compress large ranges of corruption into narrow score intervals, reducing interpretability and limiting the ability to distinguish gradual improvements or deteriorations in segmentation quality. In contrast, non-threshold-based overlap metrics, particularly MMA, produce smoother and more gradual degradation curves that better reflect the continuous nature of segmentation quality deterioration. Rather than exhibiting abrupt collapse, MMA maintains sensitivity throughout the full corruption range, allowing progressive improvements or failures to remain measurable.

Among the evaluated metrics, AP@50 demonstrates the strongest instability due to its binary TP/FP/FN formulation with fixed IoU thresholding. In $\textit{Erosion}$, $\textit{Dilation}$, and $\textit{Shifting}$, AP@50 follows a characteristic nonlinear decay pattern: scores initially remain high while most objects exceed the threshold, rapidly collapse once overlaps fall below the threshold for the majority of objects, and finally plateau as only the largest objects remain matched. This produces steep score discontinuities that poorly reflect the gradual degradation visible in the corrupted masks and highlights a fundamental limitation of binary thresholding for evaluating progressive segmentation corruption.

PQ exhibits behavior intermediate between strict threshold-based metrics and continuous overlap-based metrics. In corruption types dominated by overlap degradation, including $\textit{Erosion}$, $\textit{Dilation}$, and $\textit{Shifting}$, PQ decreases earlier than AP@50 because it incorporates segmentation quality within matched instances rather than relying solely on binary TP/FP/FN counts. Consequently, PQ remains sensitive even before overlap thresholds are crossed. However, in corruption types dominated primarily by object-level errors, such as $\textit{Removal}$, $\textit{Addition}$, and $\textit{Clumping}$, PQ degrades substantially more slowly because FP and FN penalties are weighted by one-half within the denominator. This weighting reduces responsiveness to accumulating detection errors and dampens overall score degradation.

SEG behaves differently from AP@50 and PQ due to its GT overlap-based thresholding formulation. In $\textit{Dilation}$ and $\textit{Clumping}$, SEG decreases more gradually because boundary expansion does not reduce overlap with the original GT masks. As a result, SEG behaves somewhat more similarly to overlap-based metrics under these corruption types. However, this apparent robustness arises partly from the lack of explicit FP penalization, which becomes more apparent in the FP corruption experiments discussed below.

\subsubsection{False Positive Handling and Scale Sensitivity}
The $\textit{Addition}$ and $\textit{Variable-size Addition}$ experiments reveal substantial differences in how metrics handle FP object detections. In both of these experiments, SEG remains fixed at 1.0 despite the introduction of numerous nonoverlapping objects because FP predictions are not independently penalized. A related effect is partially visible in $\textit{Fragmentation}$, where SEG decreases relatively slowly compared to PQ and AP@50 because newly created fragments contribute only in implicit penalties of decreased overlap whereas PQ and AP@50 explicitly count each new fragment as FPs.

Additionally, the $\textit{Variable-size Addition}$ experiment exposes a major weakness of threshold-based metrics in handling FP object detections. As mentioned above, SEG remains entirely unchanged because it does not penalize nonoverlapping added predictions. AP@50 and PQ do change in that they add ten more FPs to their calculation, however, they remain constant after the first iteration because they do not consider the magnitude of these predictions. It can also be explicitly seen here how PQ’s reduced weighting of FPs in its denominator dampens their impact compared to AP@50. In contrast to the threshold-based metrics, MMA continues to decrease smoothly throughout the corruption range, indicating sensitivity to increasing FP magnitude even when the number of objects remains constant. This behavior suggests that MMA better captures the effect of object misdetection, rather than responding primarily to object counts.

\subsubsection{Matching and Aggregation Behavior}
Comparison between MMA and other overlap-based metrics highlights the importance of MMA’s globally optimal matching formulation. MMA consistently achieves higher scores than MMA-Greedy in corruption settings where assignment ambiguity becomes significant, including $\textit{Dilation}$, $\textit{Clumping}$, and $\textit{Shifting}$. Under these corruption types, greedy assignment strategies can make locally optimal matches early that prevent better overall correspondences later in the matching process. MMA instead computes globally optimal assignments, allowing it to better preserve meaningful object correspondences under under-segmentation and partial overlap. The increasing separation between MMA and MMA-Greedy as corruption severity grows suggests that optimal matching becomes progressively more important as matching becomes more ambiguous.

Although AJI is also overlap-based, its response characteristics differ notably from MMA in corruption settings involving instance-level errors. In $\textit{Fragmentation}$ and $\textit{Clumping}$, AJI often decreases rapidly during early corruption stages before partially leveling out, indicating decreased stability and sensitivity at different levels of corruption. This variability likely arises from AJI’s aggregation and matching formulations that allow multi-counting of pixels and include explicit FP penalization as observed in the synthetic failure cases. MMA’s global normalization strategy maintains smoother degradation curves that more consistently track the progressive loss of object integrity. These differences suggest that MMA provides a more faithful representation of gradual segmentation degradation under complex structural corruptions.

Each evaluated metric captures certain desirable aspects of segmentation quality. AP@50 strongly emphasizes strict object detection correctness, PQ partially preserves overlap quality within matched instances, SEG is sensitive under sufficient GT overlap, and AJI and MMA-Greedy retain continuous overlap information without hard thresholding. However, these metrics also exhibit important limitations under progressive corruption, including threshold-driven discontinuities, insensitivity to FP magnitude, ambiguous overlap accounting, or suboptimal assignment behavior.

Across all evaluated corruption types, MMA demonstrates the strongest overall balance between stability and sensitivity. Unlike threshold-based metrics, MMA avoids abrupt threshold-driven collapse and preserves meaningful score resolution throughout progressive degradation. Unlike SEG, it remains sensitive to FP accumulation, and it does not exhibit severe nonlinear discontinuities such as those in AP@50. Compared to PQ, MMA maintains stronger responsiveness to object-level corruption without dampening penalties through denominator weighting. Finally, in contrast to MMA-Greedy and AJI, MMA benefits from globally optimal assignment and more stable overlap aggregation, producing smoother and more interpretable degradation trajectories under challenging matching scenarios. Collectively, these properties suggest that MMA provides a more calibrated and biologically meaningful evaluation of instance segmentation quality under realistic corruption conditions.

\subsection{Model Ranking Comparison}
With the robustness, stability, and sensitivity of MMA established in the previous two experiments, we now investigate whether MMA affects model selection decisions. To determine this, we quantify ranking disagreement between MMA and each baseline metric using two complementary measures: pairwise disagreement and top-1 disagreement. Pairwise disagreement measures how often MMA and a baseline metric differ in their rankings at a fine-grained level between model pairs, while top-1 disagreement measures whether each metric selects a different best-performing model than MMA, directly reflecting changes in final model selection.

\begin{table}[width=.9\linewidth,cols=3,pos=h]
\centering
\caption{Pairwise and top-1 ranking disagreement between MMA and baseline metrics on the LIVECell test set.}
\begin{tabular}{lcc}
\toprule
\textbf{Baseline Metric} & \textbf{Pairwise Disagreement} & \textbf{Top-1 Disagreement}\\
\midrule
AJI & 0.17 & 0.25\\
SEG & 0.25 & 0.30\\
PQ & 0.23 & 0.46\\
AP@50 & 0.27 & 0.44\\
\bottomrule
\label{TABLE:ranking_disagreement}
\end{tabular}
\end{table}

As shown in Table \ref{TABLE:ranking_disagreement}, all baseline metrics exhibit non-trivial disagreement with MMA at both levels of analysis. Pairwise disagreement ranges from 0.17 (AJI) to 0.27 (AP@50), indicating that MMA changes a substantial fraction of model rankings from their ordering in baseline evaluation metrics. Top-1 disagreement is even more pronounced, with values ranging from 0.25 (AJI) to 0.46 (PQ), demonstrating that baseline metrics can select a different best-performing model than MMA in up to half the evaluated examples. Notably, PQ and AP@50 show the highest top-1 disagreement, suggesting that threshold-based evaluation schemes differ the most from MMA in their relative scoring of model outputs.

These results show that metric choice can lead to materially different model rankings and, in some cases, different top-performing methods. The observed disagreement between MMA and baseline metrics indicates that differences in metric formulation do not merely affect absolute scores, but can directly influence conclusions drawn from instance segmentation benchmarks. In high stakes settings where benchmark outcomes guide future methodological development, this highlights the importance of using well-justified and consistent evaluation metrics for fair model comparison.

\section{Limitations}
While we show MMA to behave the most intuitively under common instance segmentation failure cases and also exhibit the strongest stability and sensitivity balance during progressive mask corruption, there exist several limitations to its formulation. Primarily, certain properties that we demonstrate MMA to possess are undesirable in some scenarios. For example, in situations where detection accuracy is more important than global image segmentation quality (e.g. cell counting), a normalization strategy that weights each object equally (SEG, PQ, AP@50) instead of weighting each pixel equally as MMA does will likely produce scores that better indicate performance on the desired task. Additionally, as seen in the $\textit{Over-Segmentation}$ example in Fig. \ref{FIG:synthetic_failure_cases}, MMA’s lack of explicit penalization for unmatched predicted objects makes it insensitive to redundant predictions that overlap with alternatively matched GT objects. A final limitation of MMA is its time complexity. MMA’s maximum matching procedure has complexity O($n^3$) while greedy matching has complexity O($n^2$), significantly increasing MMA’s evaluation time on images with high object density.

\section{Conclusion}
We introduced Maximum Matching Accuracy (MMA), a continuous, threshold-free instance segmentation metric that combines globally optimal one-to-one matching with pixel-level normalization. MMA was designed to address three common limitations of existing instance segmentation metrics: threshold-dependent scoring, non-optimal or non-exclusive matching, and object-level normalization that can distort the contribution of segmentation errors. Through synthetic failure cases, we demonstrated how these design choices can lead baseline metrics to produce discontinuous, unstable, or unintuitive evaluations under common segmentation failure modes. Across progressive corruption experiments, MMA consistently exhibited smooth score degradation while remaining sensitive to both overlap deterioration and false-positive accumulation. Together, these results demonstrate that MMA provides a more reliable and interpretable measure of segmentation quality than existing alternatives while avoiding many of the failure modes introduced by thresholding, ambiguous matching, and object-centric normalization. The effect of using MMA for model benchmarking is further highlighted in our ranking comparison experiment where we show that MMA can lead to different model selection outcomes compared to baseline metrics in up to 50\% of evaluated cases. These results indicate that MMA offers a principled and impactful framework for benchmarking instance segmentation models in biological imaging. Future work should investigate MMA’s behavior in other segmentation domains and explore extensions for tasks where alternative notions of object-level importance or matching constraints are desirable.

\section*{Code and data availability
}
The LIVECell Dataset is accessible at:  \url{https://sartorius-research.github.io/LIVECell/}. We provide relevant code and the synthetic test cases at \url{https://github.com/kadenstillwagon/MMA}.

\section*{Acknowledgement}
We gratefully acknowledge the NIH R01NS102727, NIH Single Cell Grant 1 R01 EY023173, NIH R01DA029639 and NIH RF1AG079269, support from Georgia Tech through the Institute for Bioengineering and Biosciences, Invention Studio, PACE computing infrastructure, and the George W. Woodruff School of Mechanical Engineering.

\bibliographystyle{cas-model2-names}

\bibliography{mma}

\newpage

\bio{figs/kaden_headshot}
Kaden Stillwagon is a Computer Science masters student at Georgia Tech. He holds a B.S. in Computer Science from Georgia Tech. His research focuses on the development of Machine Learning models for morphological and dynamical quantification of biological cells.
\endbio

\vskip3pc

\bio{figs/ally_headshot}
Alexandra VandeLoo is a Bioengineering doctoral student at Georgia Tech. She holds a B.S. in Materials Engineering and Biochemistry from Iowa State and an M.S. in Biomedical Engineering from Georgia Tech. Her research focuses on tool creation for quantification of qualitative biological techniques.
\endbio

\vskip3pc

\bio{figs/craig_headshot}
Craig Forest is a Professor and Regent’s Entrepreneur in the George W. Woodruff School of Mechanical Engineering at Georgia Tech where he also holds program faculty positions in Bioengineering and Biomedical Engineering as well as an adjunct appointment in the College of Computing. He conducts research on biological measurement and control using robotic instrumentation.
\endbio

\vskip3pc

\end{document}